\pdfoutput=1

\documentclass[11pt]{article}

\usepackage[final]{acl}

\usepackage{times}
\usepackage{latexsym}
\usepackage{multirow}
\usepackage{makecell} 
\usepackage{xcolor}
\usepackage{booktabs}
\usepackage{multirow}
\usepackage{xcolor}
\usepackage{graphicx}

\usepackage[T1]{fontenc}

\usepackage[utf8]{inputenc}

\usepackage{microtype}

\usepackage{inconsolata}

\usepackage{graphicx}
\usepackage{amsmath}
\usepackage{subcaption}

\title{ECO Decoding: Entropy-Based Control for Controllability and Fluency \\ 
in Controllable Dialogue Generation}

\author{Seungmin Shin \\
  SungKyunKwan University \\
  Republic of Korea \\
  \texttt{seungminshin00@gmail.com} \\\And
  Dooyoung Kim \\
  SungKyunKwan University \\
  Republic of Korea \\
  \texttt{kdysunleo98@gmail.com} \\\And
  Youngjoong Ko\thanks{Corresponding author} \\
  SungKyunKwan University \\
  Republic of Korea \\
  \texttt{yjko@skku.edu} \\}

\begin{document}
\maketitle
\begin{abstract}
Controllable Dialogue Generation (CDG) enables chatbots to generate responses with desired attributes, and weighted decoding methods have achieved significant success in the CDG task. However, using a fixed constant value to manage the bias of attribute probabilities makes it challenging to find an ideal control strength that satisfies both controllability and fluency. To address this issue, we propose \textbf{ECO decoding} (\textbf{E}ntropy-based \textbf{CO}ntrol), which dynamically adjusts the control strength at each generation step according to the model’s entropy in both the language model and attribute classifier probability distributions. Experiments on the DailyDialog and MultiWOZ datasets demonstrate that ECO decoding consistently improves controllability while maintaining fluency and grammaticality, outperforming prior decoding methods across various models and settings. Furthermore, ECO decoding alleviates probability interpolation issues in multi-attribute generation and consequently demonstrates strong performance in both single- and multi-attribute scenarios.
\end{abstract}

\section{Introduction}

In recent studies, Controllable Dialogue Generation (CDG) \cite{zhang2023semantic,zeng2023seen} has been proposed to enhance the realism and accuracy of responses generated by conversational models, improving the user experience. CDG enables chatbots to generate responses tailored to desired attributes like emotion and dialog-act.

Recently, training-based methods such as alignment tuning \cite{yang2024metaaligner} and weighted decoding approaches \cite{Yang_2021,arora2022director} have achieved notable success in the field of controllable generation. While alignment tuning requires re-training large models from scratch, weighted decoding can be applied at inference time by multiplying the attribute probabilities from a trained classifier with the language model distribution. This design enables controlled response generation with minimal additional data and training overhead, making weighted decoding an appealing strategy for CDG. Consequently, we focused on this weighted decoding strategy to effectively generate controllable responses.

In weight decoding methods, generating responses controlled by desired attributes involves the adjustment of the next token probability distribution modeled by the language model. This is achieved by multiplying the attribute probability of the generated response obtained from the attribute classifier and the next token probability. In this process, the control strength is used as the exponent of the attribute probability to control attribute bias. As the control strength increases, the generated tokens become more dependent on the token rank of attribute probability.

\begin{figure}[t!]
    \includegraphics[width=\columnwidth]{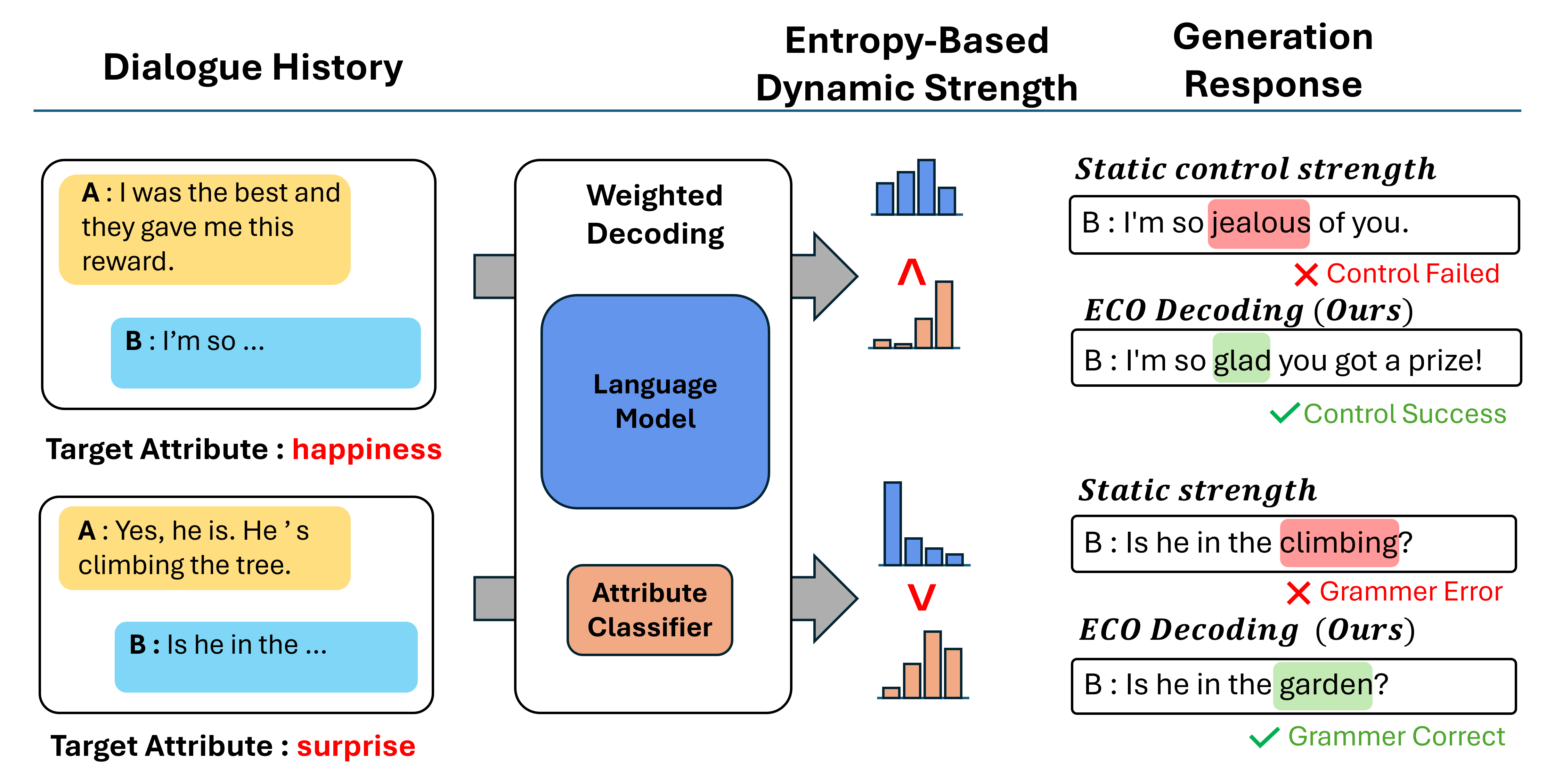}
    \caption{An example of a controllable dialogue generation method based on dynamic weighting with ECO decoding. ECO decoding dynamically adjusts the weights between the language model probability distribution and the attribute control probability distribution, enabling attribute control while maintaining fluency.}
    \label{fig:intro}
\end{figure}

Multiplying the attribute probability alters the probability distribution of the language model, which can affect language modeling performance. When static control strength is used, the same control probability is continuously reflected in the generated sentence, even if the sentence has already received sufficient attribute control or if specific words need to be generated for fluency. This can lead to a trade-off between controllability and fluency. Furthermore, the fact that the appropriate control strength varies depending on the situation is an important issue. If this is not properly accounted for, it can lead to decreased efficiency. Figure \ref{fig:intro} shows an example of a failed response generation with these fixed static control strength.

\begin{figure*}[t!]
    \centering
    \includegraphics[width=\textwidth]{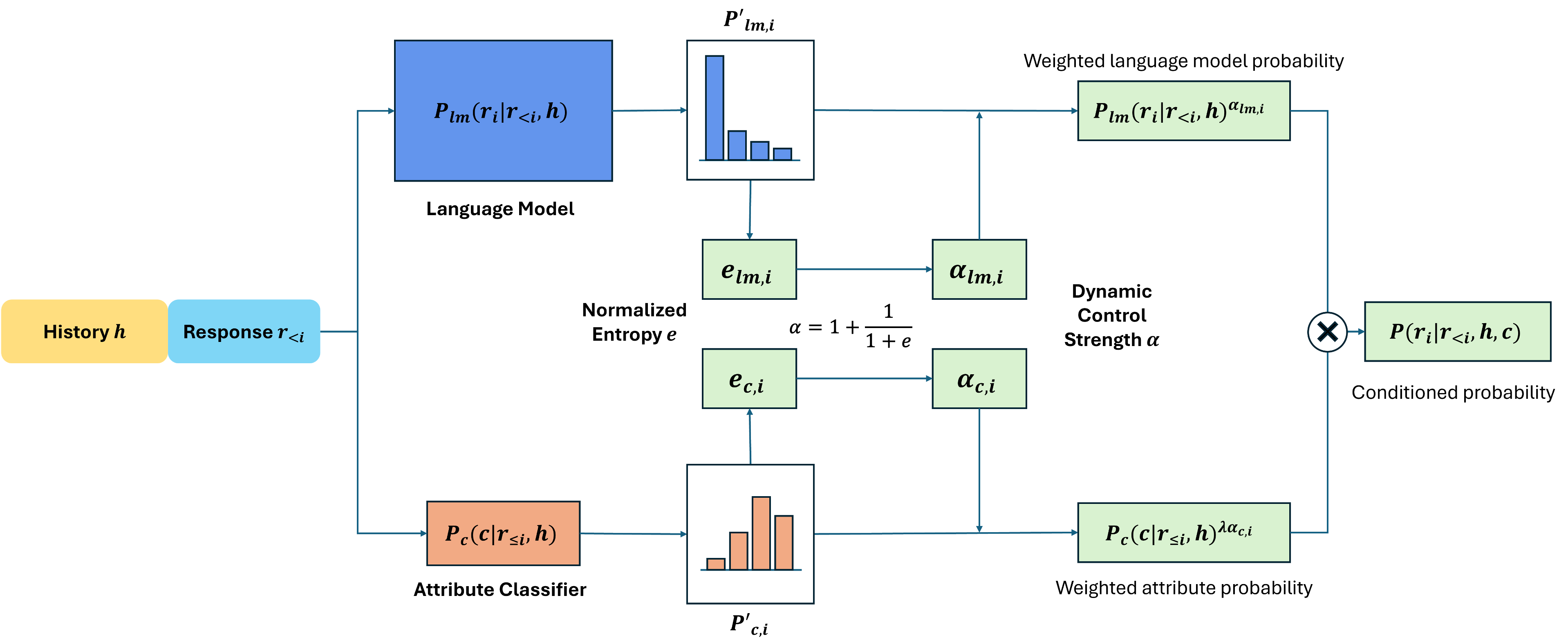}
    \caption{An illustration of controllable dialogue generation using the weighted decoding method, incorporating ECO decoding.}
    \label{fig:main}
\end{figure*}

To address this, we propose \textbf{ECO} (\textbf{E}ntropy-based \textbf{CO}ntrol strength) \textbf{Decoding}, which dynamically adjusts control strength at each decoding step. Specifically, we compute the entropy \cite{shannon2001mathematical} of both the language model’s probability distribution and the attribute classifier’s distribution at every step. If the language model assigns high probability (low entropy) to a particular token, we prioritize the model’s prediction to preserve fluency. In contrast, when the model exhibits low confidence (high entropy), we increase the relative weight of the attribute probabilities to ensure stronger attribute control. Notably, our analysis result using \emph{part-of-speech (POS)} in Appendix~\textbf{}{A.1} demonstrates that function words with fixed grammatical roles often yield lower entropy, whereas content words (e.g., nouns, verbs, adjectives) show higher entropy. This aligns with our intuition that words with higher uncertainty provide more leeway for attribute manipulation without significantly compromising fluency.

This dynamic control method effectively balances the language model's fluency with the attribute classifier's controllability, thereby achieving an optimal trade-off between naturalness and the desired attribute expression in the final generated sentences.
To validate our intuition, we experiment with three existing controllable generation models using the DailyDialog \cite{li-etal-2017-dailydialog} dataset and MultiWOZ \cite{budzianowski2018towards} dataset. Experimental results demonstrate that ECO decoding achieves high controllability while maintaining text fluency across all models.

Our main contributions are as follows:\begin{enumerate}
\item{We raise the issue of static control strength in existing weighted decoding methods and propose a dynamic control strength approach to generate responses with high controllability as well as maintain fluency.} 
\item{We show that ECO decoding can be generalized to multi-attribute scenarios, alleviating probability interpolation issues that are commonly  to single-attribute weighted decoding methods.} 
\item{Experimental results show that the ECO decoding method outperforms the existing weighted decoding methods for all existing controllable generation models.} 
\end{enumerate}

\section{Related Work}
Previous approaches such as GeDi \cite{krause2021gedi} and DExperts \cite{liu2021dexperts} adopt weighted decoding, but they require a separate language model for each attribute value. In multi-attribute settings, this design is inefficient, as multiple models must be loaded during inference, leading to increased memory usage and latency. While ECO decoding can in principle be combined with these methods, in this work we focus on three baselines—FUDGE, Director, and DASC—that are more scalable and practically applicable to complex controllable generation scenarios.
\subsection{Weighted Decoding}
Controllable dialogue generation aims to generate a response, $R = \{r_1, r_2,...,r_N\}$, with desired attributes, given dialogue history \(h\) and attribute \(c\), using a pre-trained auto-regressive model (e.g. GPT2, \cite{radford2019language}, DialoGPT \cite{zhang2020DialoGPT}). Emotion and dialog-act can be attributes for controllable dialogue generation.

To condition on attribute c, the response generation given a dialogue can be formulated as follows:

\vspace{-2mm}
\begin{equation}
P(R|h, c) = \prod_{i=1}^{N}P(r_i|r_{<i}, h, c)
\end{equation}
Using Bayesian factorization, $P(r_i|r_{<i}, h, c)$ can be converted into the following equation.

\vspace{-2mm}
\begin{equation}
P(r_i|r_{<i}, h, c) \propto P(r_i|r_{<i},h)P(c|r_{\leq i},h)^\lambda
\end{equation}
\noindent where the first term $P(r_i|r_{<i},h)$ represents the next token probability modeled by a language model, and the second term $P(c|r_{\leq i},h)$ represents the  attribute probability of the generated response obtained from the attribute
classifier.
In addition, control strength $\lambda$ is added to the exponential term of the attribute probabilities to control attribute bias.

When dealing with multi-attribute control, Equation \ref{eq:multi-attr-base} can be extended by introducing the product of multiple attribute classifiers, assuming that the attributes are conditionally independent:

\vspace{-2mm}
\begin{equation}
\label{eq:multi-attr-base}
P(r_i|r_{<i}, h, C) \propto P(r_i|r_{<i},h)\prod_{c_{j} \in C} P(c_{j}|r_{\leq i},h)^\lambda
\end{equation}

where $C$ denotes the set of target attributes. The product of probabilities is typically implemented as the sum of logits.

\subsection{Weighted Decoding Models}
\paragraph{FUDGE} \citealp{Yang_2021} trained a classification model for partial sequences through an external attribute classifier. Specifically, for each training example $\{(x,c)\}$, where $x$ is sentence and $c$ is class label, the classifier is trained on all partial sequences $\{(x_{1:i},c)\}$ at each step. During inference, at a given time step $i$, the classifier predicts the probability that appending the top $k$ candidate tokens to the generated text will satisfy the attribute $c$ in future generations. 

\paragraph{Director} \citealp{arora2022director} addressed the inefficiency issue of requiring a external model during inference. It integrates the language model and attribute classification functionality into a single model, overcoming the inefficiency of the external classifier evaluating the attribute for every candidate token. To address this issue, an additional classification head is introduced, which takes the last hidden state as input and computes the probability that each token in the vocabulary satisfies the specified attribute. This allows for the effective incorporation of attribute information without the need for a external classifier.

\paragraph{DASC} \citealp{zhang2023semantic} addressed the computational inefficiency issues arising from dual-head architectures. DASC introduces Attribute Token Embedding and Attribute Semantic Embedding concepts, employing a semantic space-based weighted decoding mechanism to reduce the number of parameters while improving computational efficiency. Each token is associated with an embedding that captures its attribute semantics, and these embeddings are projected into an attribute semantic space via attribute-specific linear layers. This design facilitates smooth control over multiple attributes and enables effective interpolation among attribute embeddings, allowing more diverse range of attribute combinations.

\section{Methodology}
\subsection{Entropy-based Control Strength} 

The existing weighted decoding methods apply a fixed control strength and they are not flexible enough to handle situations where stronger or no more control is needed.
In such cases, they may fail to control attribute, or even if they succeed, the fluency and grammar may be degraded.
To solve this problem, we propose the ECO decoding method that utilizes the entropy of the probability distribution to dynamically adjust the control strength.
Dynamic control strength allows to achieve higher attribute control rates, while maintaining generation quality, including context and grammar.

Entropy is a measure of the uncertainty of a probability distribution, which is lower when the probability distribution is focused on a specific value and higher when it is more evenly distributed. Given this property, the higher the entropy of the next token probability distribution is, the more likely it is to contain a variety of plausible candidates. This is an advantageous property for exploring plausible options that satisfy desired attribute. Based on this insight, a novel mechanism of dynamically controlling strength is developed by weighting probability distributions from language models and it controls each property inversely to their entropy score. That is, distributions with lower uncertainty are more strongly reflected.
Figure \ref{fig:main} shows how ECO decoding is working by using dynamic control strength based on both of the language model entropy and the attribute entropy. ECO decoding can be applied to the existing weighted decoding methods and requires no additional modules or training.

\paragraph{Language Model Entropy}
Dynamic control strength $\alpha_{x,i}$ is separately calculated for $i$-th generation step, and it can have different values while a sentence is generated. To calculate control strength, we select the top-$k$ candidate tokens. From the probability distribution $P_{lm, i}$ of the language model, we construct the set $S$, which consists of the $k$ tokens with the highest probabilities. Let $P'_{lm, i}$ denote the partial probability distribution of top-$k$ tokens in $S$.

\vspace{-2mm}
\begin{equation}
{P'}_{lm,i} = \{P_{lm}(t|r_{< i}, h) | t \in S\}
\end{equation}

To convert the partial probability distribution ${P'}_{lm,i}$ into a probability distribution, we recompute the probability distribution of the top-$k$ tokens using a softmax function with temperature $\tau_{lm}$.

\vspace{-2mm}
\begin{equation}
e_{lm, i} = Entropy(Softmax({P'}_{lm, i} / \tau_{lm}))
\end{equation}

\paragraph{Attribute Entropy}
Weighted decoding methodologies for CDG utilize attribute classifier $P_{c}$ to reflect attributes. For each candidate token $t$ in the top-$k$ token set $S$, concatenates the current sequence $r_{<i}$ with $t$ and computes the probability $P_{c, i}([r_{< i}; t], h)$ which represents the probability of token $t$ being part of the generated response while aligning with the target attribute to be controlled. 
The set ${P'}_{c, i}$ is the probabilities of the target attribute for all candidate tokens in top-$k$ token set $S$. The attribute entropy $e_{c, i}$ is computed based on a probability distribution normalized by softmax the set of attribute probabilities ${P'}_{c, i}$ over $\tau_{c}$, where $\tau_{c}$ is the attribute temperature for softmax.

\paragraph{Entropy Based Control Strength}
To assign higher weights to probability distributions with higher entropy, we utilize a control strength formula with an inverse function structure, as shown in Equation \ref{eq:strength}. The control strength $\alpha_{x,i}$ is applied to both the language model probability distribution $P_{lm}$ and the attribute probability distribution $P_{c}$.
The language model probability distribution and the attribute probability distribution are reflected by a power of their respective weight $\alpha_{x,i}$. The attribute probability distribution additionally reflects the strength scale factor $\lambda$. The value of $\lambda$ allows to adjust whether to focus more on attribute control or language modeling performance. The final probability distribution for generating the next token $P(r_i|r_{<i}, h, c)$ is computed by multiplying the two weighted probability distributions as shown in Equation \ref{eq:single-attr}. If each of the control strength $alpha$ values were fixed at 1, the same result would be obtained as with the traditional weighted decoding methodologies.

\vspace{-2mm}
\begin{equation}
\label{eq:strength}
\alpha_{x,i} = 1 + (\frac{1}{1+e_{x,i}})
\end{equation}

\vspace{-2mm}
\begin{equation}
\begin{split}
\label{eq:single-attr}
P(r_i|r_{<i}, h, c) & \propto P_{lm}(r_i|r_{<i},h)^{\alpha_{lm,i}} \\
& \times P_{c}(c|r_{\leq i},h)^{\lambda *{\alpha_{c, i}}}
\end{split}
\end{equation}

\subsection{Multiple Attribute Control Strength} 
Existing weighted decoding methodologies struggle to control multiple attributes simultaneously due to their fixed control strength. 
When using a fixed control strength for each attribute, the search space of attribute control strengths grows exponentially. Furthermore, even when control strength is applied, effectively incorporating more than two attributes remains a main challenge. In contrast, our proposed ECO-decoding method enables CDG to control generation by reformulating the final probability distribution based on multiple attributes. Dynamic control strength $\alpha_{x, i}$ adjusts the weight of probability distributions at each generation step based on the entropy of the language model and the entropy of each attribute, allowing more flexible and adaptive multi-attribute control. When $C$ is the set of controlling attributes, the multiple attribute control formula for ECO-decoding is as follows:

\begin{equation}
\begin{aligned}
\label{eq:multi-attr}
P(r_i|r_{<i}, h, C) 
&\propto P_{lm}(r_i|r_{<i},h)^{\alpha_{lm, i}} \\
&\hspace{-1.5cm} \times \prod_{c_{j} \in C} P_{c_{j}}(c_{j}|r_{\leq i},h)^{\lambda *{\alpha_{c_{j}, i}}}
\end{aligned}
\end{equation}

\vspace{-2mm}
\begin{equation}
{P'}_{c, i} = \{P_{c}([r_{< i}; t], h) | t \in S\}
\end{equation}

\vspace{-2mm}
\begin{equation}
e_{c, i} = Entropy(Softmax({P'}_{c, i}/\tau_{c}))
\end{equation}

\begin{table*}[t!]
\centering
{
\scalebox{0.95}{ 
\begin{tabular}{lllllll}
\hline
\multicolumn{1}{c}{Method} & \multicolumn{1}{c}{Accuracy} & \multicolumn{1}{c}{Rouge-1} & \multicolumn{1}{c}{Rouge-L} & \multicolumn{1}{c}{Dist-1} & \multicolumn{1}{c}{Dist-2} & \multicolumn{1}{c}{Grammar} \\ \hline \hline
\multicolumn{7}{c}{Emotion}                                                                                                                                                                                  \\ \hline
DialoGPT(176M)                  & \multicolumn{1}{c}{-}        & 9.00       & 8.53       & 0.58                       & 0.76                       & 90.21                       \\ \hline
FUDGE                    & 76.98                        & 9.06                        & 8.60                        & 0.60                       & 0.75                       & 90.30                       \\
+ ECO decoding        & 81.03 \scriptsize{\textcolor{red}{(+4.05)}}                       & 9.13 \scriptsize{\textcolor{red}{(+0.07)}}                       & 8.64 \scriptsize{\textcolor{red}{(+0.04)}}                        & 0.62 \scriptsize{\textcolor{red}{(+0.02)}}                        & 0.75 \scriptsize{(-)}                        & 90.34 \scriptsize{\textcolor{red}{(+0.04)}}                        \\\hline
Director                  & 79.94                        & 8.83                        & 8.37                        & 0.59                       & 0.70                       & 90.23                       \\
+ ECO decoding       & 82.82 \scriptsize{\textcolor{red}{(+2.88)}}                        & 8.82 \scriptsize{\textcolor{blue}{(-0.01)}}                       & 8.34 \scriptsize{\textcolor{blue}{(-0.03)}}                        & 0.59 \scriptsize{(-)}                       & 0.71 \scriptsize{\textcolor{red}{(+0.01)}}                       & 90.30 \scriptsize{\textcolor{red}{(+0.07)}}                       \\\hline
DASC                      & 74.65                        & 8.25                        & 7.87                        & 0.58                       & 0.70                       & 90.30                       \\
+ ECO decoding       & 75.74 \scriptsize{\textcolor{red}{(+1.09)}}                        & 8.22 \scriptsize{\textcolor{blue}{(-0.03)}}                       & 7.79 \scriptsize{\textcolor{blue}{(-0.08)}}                        & 0.58 \scriptsize{(-)}                        & 0.71 \scriptsize{\textcolor{red}{(+0.01)}}                       & 90.39 \scriptsize{\textcolor{red}{(+0.09)}}  \\\hline
\multicolumn{7}{c}{Dialog-act}                                                                                                                                                                               \\ \hline
DialoGPT(176M)                 & \multicolumn{1}{c}{-}        & 9.14       & 8.66       & 0.57                       & 0.78                       & 91.24                       \\ \hline
FUDGE                    & 41.07                        & 9.21                        & 8.75                        & 0.59                       & 0.78                       & 90.98                             \\
+ ECO decoding            & 46.42 \scriptsize{\textcolor{red}{(+5.35)}}                        & 9.21 \scriptsize{(-)}                       & 8.79 \scriptsize{\textcolor{red}{(+0.04)}}                       & 0.62 \scriptsize{\textcolor{red}{(+0.03)}}                       & 0.79 \scriptsize{\textcolor{red}{(+0.01)}}                       & 91.00 \scriptsize{\textcolor{red}{(+0.02)}}                       \\\hline
Director                  & 70.96                        & 10.43                       & 9.94                        & 0.62                       & 0.78                       & 91.18                             \\
+ ECO decoding            & 71.56 \scriptsize{\textcolor{red}{(+0.60)}}                        & 10.46 \scriptsize{\textcolor{red}{(+0.03)}}                       & 9.96 \scriptsize{\textcolor{red}{(+0.02)}}                       & 0.63 \scriptsize{\textcolor{red}{(+0.01)}}                       & 0.79 \scriptsize{\textcolor{red}{(+0.01)}}                       & 91.15 \scriptsize{\textcolor{blue}{(-0.03)}}                       \\\hline
DASC                      & 42.59                        & 9.53                        & 9.03                        & 0.59                       & 0.75                       & 91.13                             \\
+ ECO decoding            & 47.17 \scriptsize{\textcolor{red}{(+4.58)}}                        & 9.52 \scriptsize{\textcolor{blue}{(-0.01)}}                       & 9.05 \scriptsize{\textcolor{red}{(+0.02)}}                       & 0.60 \scriptsize{\textcolor{red}{(+0.01)}}                       & 0.76 \scriptsize{\textcolor{red}{(+0.01)}}                       & 91.13 \scriptsize{{(-)}}                             \\ \hline
\end{tabular}
}
}
\caption{Evaluation results for a single attribute of emotion or dialog-act on the DailyDialog test set. The scores in parentheses indicate the performance gap between static and dynamic control settings.}
\label{tab:main_result}
\end{table*}

\begin{table*}[t!]
\centering
{
\resizebox{\textwidth}{!}{ 
\begin{tabular}{llllllll}
\hline
\multicolumn{1}{c}{Method} & \multicolumn{1}{c}{Accuracy(Emo)} & \multicolumn{1}{c}{Accuracy(Act)} & \multicolumn{1}{c}{Rouge-1} & \multicolumn{1}{c}{Rouge-L} & \multicolumn{1}{c}{Dist-1} & \multicolumn{1}{c}{Dist-2} & \multicolumn{1}{c}{Grammar} \\ \hline \hline
DialoGPT(176M) & \multicolumn{1}{c}{-} & \multicolumn{1}{c}{-} & 9.00 & 8.53 & 0.58 & 0.76 & 90.21 \\ \hline
FUDGE    & 66.17 & 44.17 & 8.21 & 7.82 & 0.57 & 0.74 & 90.20 \\
+ ECO decoding & 66.41 \scriptsize{\textcolor{red}{(+0.24)}} & 45.57 \scriptsize{\textcolor{red}{(+1.40)}}                & 8.20 \scriptsize{\textcolor{blue}{(-0.01)}} & 7.81 \scriptsize{\textcolor{blue}{(-0.01)}}                & 0.58 \scriptsize{\textcolor{red}{(+0.01)}} & 0.74 \scriptsize{(-)}                  & 90.21 \scriptsize{\textcolor{red}{(+0.01)}}     \\\hline
Director & 80.48 & 60.65 & 9.41 & 8.99 & 0.58 & 0.73 & 90.22 \\
+ ECO decoding & 81.18 \scriptsize{\textcolor{red}{(+0.7)}} & 61.20 \scriptsize{\textcolor{red}{(+0.65)}}    & 9.49 \scriptsize{\textcolor{red}{(+0.08)}} & 8.97 \scriptsize{\textcolor{blue}{(-0.02)}}                  & 0.58 \scriptsize{(-)} & 0.74 \scriptsize{\textcolor{red}{(+0.01)}} & 90.23 \scriptsize{\textcolor{red}{(+0.01)}}     \\\hline
DASC           & 75.19 & 51.17 & 8.22 & 7.67 & 0.60 & 0.77 & 90.05 \\
+ ECO decoding & 77.22 \scriptsize{\textcolor{red}{(+2.03)}} & 54.12 \scriptsize{\textcolor{red}{(+2.95)}}                & 7.60 \scriptsize{\textcolor{blue}{(-0.62)}} & 7.15 \scriptsize{\textcolor{blue}{(-0.52)}}                & 0.61 \scriptsize{\textcolor{red}{(+0.01)}}  & 0.78 \scriptsize{\textcolor{red}{(+0.01)}} & 90.19 \scriptsize{\textcolor{red}{(+0.14)}}  \\\hline
\end{tabular}
}
}
\caption{Evaluation results for multiple attributes setting on the DailyDialog test set.}
\label{tab:main_multi}
\end{table*}

\section{Experiments}
\subsection{Datasets}
\textbf{DailyDialog} \cite{li-etal-2017-dailydialog} is an English open-domain dialogue dataset containing two main attributes (emotion, dialog-act). We treat each utterance as a response and its preceding utterances as the dialogue history. The dataset provides four classes for dialog-act (inform, question, directive, commissive) and six classes for emotion (anger, disgust, fear, happiness, sadness, surprise), excluding “no emotion.” It comprises 13,681, 882, and 1,286 examples for training, validation, and test, respectively.

\textbf{MultiWOZ} \cite{budzianowski2018towards}  is a large-scale multi-domain dialogue dataset constructed from real human-to-human conversations, encompassing multi-turn dialogues across seven domains (restaurants, trains, attractions, hotels, taxis, hospitals, and police). Since the dataset itself does not provide labeled attributes, we employ an attribute classifier—used for evaluation—to label each utterance, and use these labels as the ground truth.

\subsection{Experimental Settings}
\paragraph{Language Model}
We employ DialoGPT \cite{zhang2020DialoGPT}, pre-trained on a large-scale dialogue corpus, as our baseline. Most experiments use DialoGPT-small (176M), and we also evaluate DialoGPT-large (1.1B) for scalability. Additionally, we use Llama2-7B \cite{touvron2023llama} to test the general applicability of ECO decoding to large language models.

\paragraph{Weighted Decoding Methods}
We evaluate and compare the performances of ECO decoding with those of various controllable generation models with weighted decoding methods, including FUDGE \cite{Yang_2021}, Director \cite{arora2022director}, and DASC \cite{zhang2023semantic}.

\

\paragraph{Implementation Details}
For the three weighted decoding method, the language model is frozen and each attribute classifier is trained on the training dataset. FUDGE is trained for 30 epochs with a batch size of 8 and a learning rate of 2e-5 for each attribute. For the Director, each attribute is fine-tuned for 20 epochs with a batch size of 32 and a learning rate of 1e-5. For DASC, each attribute is fine-tuned for 30 epochs with a batch size of 4 and a learning rate of 1e-5.
All methods use greedy search \cite{li2016deep}, and the maximum sequence length is set to 128. All experiments are run on a single NVIDIA GeForce RTX 3090.

Our code is available at \url{https://github.com/seummin/ECODecoding}

\subsection{Evaluation Metrics}
\paragraph{Automatic Evaluation}
To assess \emph{controllability}, we train two RoBERTa-based evaluators \cite{liu2019roberta} on the DailyDialog training set: one for emotion and one for dialog-act. These evaluators achieve 89.66\% and 80.60\% accuracy on their test sets respectively and they are used to classify generated responses. Note that these evaluators are independent from the attribute classifiers in each weighted decoding method. 
For \emph{quality}, ROUGE-1 and ROUGE-L \cite{lin2004rouge} are measured by comparing generated responses to reference answers and Dist-1 and Dist-2 \cite{li2016diversitypromoting} are computed to evaluate diversity in the generated text. For \emph{grammar}, the probability of grammaticality is utilized by the RoBERTa-based CoLA grammaticality model \cite{liu2019roberta,warstadt2019neural, morris2020textattack}.

\paragraph{Human Evaluation}
Experiments on the Director model, which showed the best performance in emotion and dialog-act attributes, conducted human evaluation based on sampling 10 contexts for each attribute value from the test set. We evaluate our generated responses based on three aspects: (1) Accuracy: Response is generated according to the desired attribute. (2) Interest: Response is specific and creative enough to capture the user's attention, and avoids repetitive or generic outputs that may reduce engagement (e.g., repeatedly generating "That's great!" for the happy attribute). (3) Sensible: Response is grammatically correct and contextually coherent.  We asked three expert evaluators to rate each metric on a scale of 1 to 5, with higher scores being better.


\begin{table*}[t!]
\centering
{
\scalebox{0.95}{ 
\begin{tabular}{lllllll}
\hline
\multicolumn{1}{c}{Model} & \multicolumn{1}{c}{Accuracy} & \multicolumn{1}{c}{Rouge-1} & \multicolumn{1}{c}{Rouge-L} & \multicolumn{1}{c}{Dist-1} & \multicolumn{1}{c}{Dist-2} & \multicolumn{1}{c}{Grammar} \\ \hline \hline
DialoGPT(176M)                  & \multicolumn{1}{c}{-}        & 9.00       & 8.53       & 0.58                       & 0.76                       & 90.21                        \\ \hline
Director                  & 79.94                        & 8.83                        & 8.37                        & 0.59                       & 0.70                       & 90.23                       \\
+ ECO decoding       & 82.82 \scriptsize{\textcolor{red}{(+2.88)}}                        & 8.82 \scriptsize{\textcolor{blue}{(-0.01)}}                       & 8.34 \scriptsize{\textcolor{blue}{(-0.03)}}                        & 0.59 \scriptsize{(-)}                       & 0.71 \scriptsize{\textcolor{red}{(+0.01)}}                       & 90.30 \scriptsize{\textcolor{red}{(+0.07)}}                       \\\hline
DialoGPT(1.1B)                  & \multicolumn{1}{c}{-}        & 11.54 & 10.89 & 0.75 & 0.73 & 87.28                 \\ \hline
Director                  & 75.66 & 11.76 & 11.15 & 0.74 & 0.73 & 87.18               \\
+ ECO decoding       & 76.05 \scriptsize{\textcolor{red}{(+0.39)}}                        & 11.82 \scriptsize{\textcolor{red}{(+0.06)}}               & 11.23 \scriptsize{\textcolor{red}{(+0.08)}}                        & 0.74 \scriptsize{(-)}                       & 0.73 \scriptsize{(-)}                       & 87.25 \scriptsize{\textcolor{red}{(+0.07)}}                       \\\hline

Llama2(7B)                  & \multicolumn{1}{c}{-}        & 15.23 & 12.99 & 0.35 & 0.08 & 90.60                 \\ \hline
Director                  & 75.43                        &15.95                         &13.74                         & 0.35                       &0.80                        &90.51                        \\
+ ECO decoding       & 75.66 \scriptsize{\textcolor{red}{(+0.23)}}                        &15.88  \scriptsize{\textcolor{blue}{(-0.07)}}                       & 13.63 \scriptsize{\textcolor{blue}{(-0.03)}}                        &  0.35\scriptsize{(-)}                       & 0.80 \scriptsize{(-)}                       & 90.55 \scriptsize{\textcolor{red}{(+0.04)}}                       \\\hline 
\end{tabular}
}
}
\caption{Evaluation results for attributes of emotion on the DailyDialog test set with various size of model.}
\label{tab:main_result_large}
\end{table*}

\subsection{Experimental Results}
\paragraph{Single Attribute Control}
Table \ref{tab:main_result} summarizes how effectively ECO decoding enhances controllability while maintaining fluency. We first run each model without attribute control to establish a baseline grammar score and then tune the control strength $\lambda$ in each method (FUDGE, Director, DASC) until we match that baseline grammar score. Eventually, we evaluate the resulting Accuracy, Dist, and ROUGE metrics. 

The results show that ECO decoding consistently improves the accuracy of both emotion and dialog-act attributes, surpassing static control methods. Notably, these improvements come without degrading grammar quality, unlike existing approaches. Furthermore, ECO decoding retains the Dist and ROUGE scores or it even slightly improves in some cases; we think that it confirms ECO decoding's ability to dynamically incorporate desired attributes without compromising fluency or overall response quality.

\begin{table*}[!t]
\centering
\renewcommand{\arraystretch}{1.05}
\resizebox{\linewidth}{!}{
\begin{tabular}{lllllllll} 
\toprule
\multicolumn{1}{c}{Model} & \multicolumn{1}{c}{Method} &
\multicolumn{1}{c}{Accuracy(Emo)} & \multicolumn{1}{c}{Accuracy(Act)} &
\multicolumn{1}{c}{ROUGE-1} & \multicolumn{1}{c}{ROUGE-2} &
\multicolumn{1}{c}{Dist-1}  & \multicolumn{1}{c}{Dist-2}  &
\multicolumn{1}{c}{Grammar} \\
\midrule
\midrule
\multirow{6}{*}{DialoGPT(176M)} 
& No attribute         & \multicolumn{1}{c}{--} & \multicolumn{1}{c}{--} & 6.72 & 6.19 & 0.53 & 0.44 & 86.42 \\ \cline{2-9}
& Director(Emo)         & 73.99 & \multicolumn{1}{c}{--} & 6.42 & 5.90 & 0.57 & 0.44 & 86.36 \\
& + ECO decoding   & 75.31 {\scriptsize\textcolor{red}{(+1.32)}} & \multicolumn{1}{c}{--} &
  6.54 {\scriptsize\textcolor{red}{(+0.12)}} &
  6.04 {\scriptsize\textcolor{red}{(+0.14)}} &
  0.55 {\scriptsize\textcolor{blue}{(-0.02)}} &
  0.44 {\scriptsize(--)} &
  86.38 {\scriptsize\textcolor{red}{(+0.02)}} \\
& Director(Act)         & \multicolumn{1}{c}{--} & 67.30 & 5.84 & 5.41 & 0.62 & 0.57 & 86.47 \\
& + ECO decoding   & \multicolumn{1}{c}{--} & 67.92 {\scriptsize\textcolor{red}{(+0.62)}} &
  5.88 {\scriptsize\textcolor{red}{(+0.04)}} &
  5.45 {\scriptsize\textcolor{red}{(+0.04)}} &
  0.63 {\scriptsize\textcolor{red}{(+0.01)}} &
  0.59 {\scriptsize\textcolor{red}{(+0.02)}} &
  86.43 {\scriptsize\textcolor{blue}{(-0.04)}} \\
& Director(Multi)       & 63.62 & 49.29 & 7.22 & 6.72 & 0.44 & 0.48 & 86.38 \\
& + ECO decoding & 66.52 {\scriptsize\textcolor{red}{(+2.90)}} &
  52.59 {\scriptsize\textcolor{red}{(+3.30)}} &
  7.35 {\scriptsize\textcolor{red}{(+0.13)}} &
  6.72 {\scriptsize(--)} &
  0.45 {\scriptsize\textcolor{red}{(+0.01)}} &
  0.49 {\scriptsize\textcolor{red}{(+0.01)}} &
  86.41 {\scriptsize\textcolor{red}{(+0.03)}} \\
\midrule
\multirow{6}{*}{Llama2(7B)}
& No attribute         & \multicolumn{1}{c}{--} & \multicolumn{1}{c}{--} & 9.84 & 7.83 & 0.22 & 0.60 & 86.91 \\ \cline{2-9}
& Director(Emo)         & 65.73 & \multicolumn{1}{c}{--} & 9.80 & 7.90 & 0.21 & 0.59 & 86.85 \\
& + ECO decoding   & 68.45 {\scriptsize\textcolor{red}{(+2.72)}} & \multicolumn{1}{c}{--} &
  10.13 {\scriptsize\textcolor{red}{(+0.33)}} &
  8.09  {\scriptsize\textcolor{red}{(+0.19)}} &
  0.20  {\scriptsize\textcolor{blue}{(-0.01)}} &
  0.58  {\scriptsize\textcolor{blue}{(-0.01)}} &
  86.85 {\scriptsize\textcolor{red}{(+0.00)}} \\
& Director(Act)         & \multicolumn{1}{c}{--} & 40.89 & 8.75 & 6.95 & 0.23 & 0.61 & 87.03 \\
& + ECO decoding   & \multicolumn{1}{c}{--} & 42.53 {\scriptsize\textcolor{red}{(+1.64)}} &
  8.98 {\scriptsize\textcolor{red}{(+0.23)}} &
  7.23 {\scriptsize\textcolor{red}{(+0.28)}} &
  0.24 {\scriptsize\textcolor{red}{(+0.01)}} &
  0.61 {\scriptsize(--)} &
  86.89 {\scriptsize\textcolor{blue}{(-0.14)}} \\
& Director(Multi)       & 59.57 & 23.90 & 6.81 & 5.65 & 0.19 & 0.58 & 86.90 \\
& + ECO decoding & 65.80 {\scriptsize\textcolor{red}{(+6.23)}} &
  28.87 {\scriptsize\textcolor{red}{(+4.97)}} &
  6.66 {\scriptsize\textcolor{blue}{(-0.15)}} &
  5.64 {\scriptsize\textcolor{blue}{(-0.01)}} &
  0.19 {\scriptsize(--)} &
  0.55 {\scriptsize\textcolor{blue}{(-0.03)}} &
  86.88 {\scriptsize\textcolor{blue}{(-0.02)}} \\
\bottomrule
\end{tabular}
} 
\caption{Evaluation results on the MultiWOZ dataset using DialoGPT (176M) and Llama2 (7B) for single-attribute settings (emotion, dialog-act) and multi-attribute settings. In the method notation, the term in parentheses specifies the attribute being controlled. The scores in parentheses indicate the performance gap between static and dynamic control settings.}
\label{tab:multiwoz_final}
\end{table*}

\begin{table}[]
\centering
\resizebox{0.95\columnwidth}{!}{%

\begin{tabular}{llll}
\hline
\multicolumn{1}{c}{Model}          & \multicolumn{1}{c}{Accuracy} & \multicolumn{1}{c}{Interest} & \multicolumn{1}{c}{Sensible} \\ \hline \hline
\multicolumn{4}{c}{Emotion}                                                                                  \\ \hline
Director       & 2.82                            & 2.96                             & 2.75                           \\
+ ECO decoding & 3.19 \textcolor{red}{(+0.37)}                            & 3.16 \textcolor{red}{(+0.20)}                             & 3.15 \textcolor{red}{(+0.40)}                           \\ \hline
\multicolumn{4}{c}{Dialog-act}                                                                               \\ \hline
Director       & 3.04                             & 2.93                             & 2.78                             \\
+ ECO decoding & 3.42 \textcolor{red}{(+0.38)}                           & 3.41 \textcolor{red}{(+0.48)}                             & 3.36 \textcolor{red}{(+0.58)}     \\ \hline             
 
\end{tabular}
}
\caption{Human Evaluation on DailyDailog test set (single attribute)}
\label{tab:human_evaluation}
\end{table}

\paragraph{Multi Attribute Control}
Multi attribute control typically involves combining attribute probabilities via multiplication. Consequently, when interpolating across multiple distributions, differences in scale and calibration can make it difficult to maintain a proper balance, often leading to a decline in overall controllability compared to single-attribute control.

In Table  \ref{tab:main_multi}, multi attribute control for the emotion attribute achieves grammar performance on par with single-attribute control, yet exhibits a decrease in overall controllability. Conversely, multi-attribute control for the dialog-act attribute appears to yield higher controllability relative to single attribute control. However, this does not necessarily indicate an actual improvement in controllability; rather, it likely reflects the selection of a relatively lower grammar score baseline due to differences in the experimental data.

By applying ECO decoding, we mitigate these interpolation issues and significantly enhance the controllability of both emotion and dialog-act attributes. Moreover, as in the single-attribute setting, dynamic weighting consistently maintains and even strengthens grammatical fluency and response quality in multi-attribute generation.

\paragraph{Language Model Scaling}
Table \ref{tab:main_result_large} compares the performance of the Director method and ECO decoding across models of varying sizes. From the smallest 176M parameter model to the 7B model, ECO decoding consistently achieves higher grammar scores while maintaining strong attribute controllability. This indicates that ECO decoding can be applied effectively to traditional weighted decoding methods regardless of model scale.

\paragraph{Generalization across Domains}
To evaluate the generalization capability of ECO decoding in diverse dialogue domains, we performed additional experiments on the MultiWOZ dataset, applying the Director Method to DialoGPT and Llama2-7B. As summarized in Table~\ref{tab:multiwoz_final}, ECO decoding improved controllability in both single- and multi-attribute settings on MultiWOZ. Importantly, these improvements were achieved without any degradation in ROUGE, grammaticality, or diversity metrics, demonstrating that ECO decoding effectively generalizes to multi-domain dialogue environments such as MultiWOZ.

\paragraph{Human Evaluation}
Table~\ref{tab:human_evaluation} presents the human evaluation results. Consistent with the automatic metrics, ECO decoding substantially outperforms the baseline in generating coherent and attribute-aligned responses. Across both emotion and dialog-act attributes, Accuracy improves by about +0.38, Interest by +0.34, and Sensible by +0.49. These results suggest that ECO decoding not only better aligns generated responses with the desired attributes but also makes them more engaging and contextually coherent from a human perspective.

\begin{figure}[t]
\vspace{-1mm}
    \centering
    \includegraphics[width=0.235\textwidth]{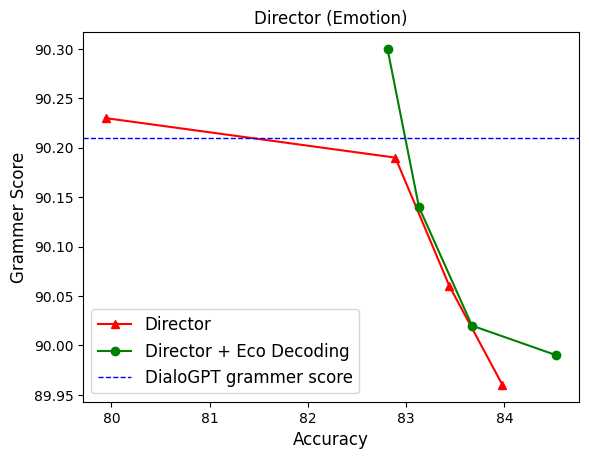}
    \includegraphics[width=.235\textwidth]{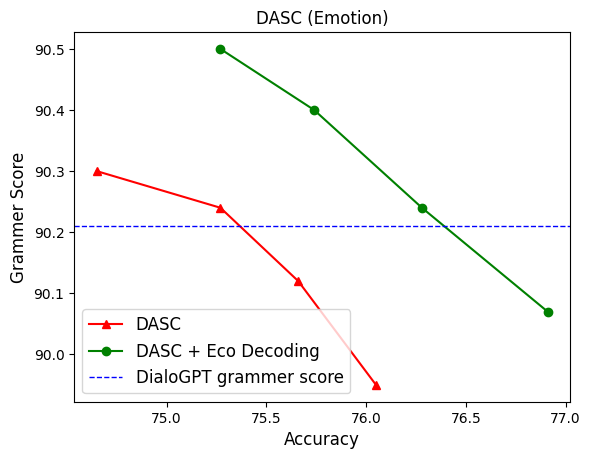}

    \textbf{\includegraphics[width=.235\textwidth]{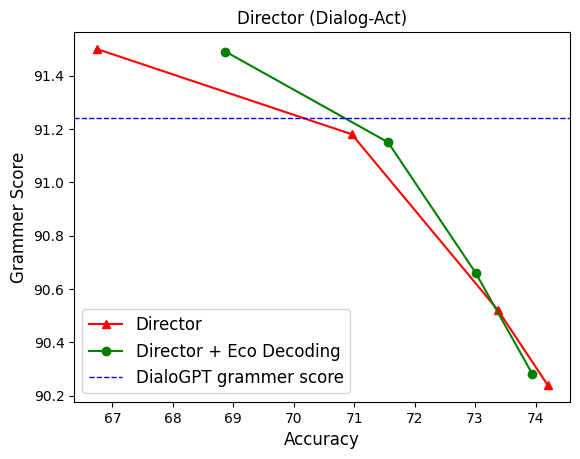}}
    \includegraphics[width=.235\textwidth]{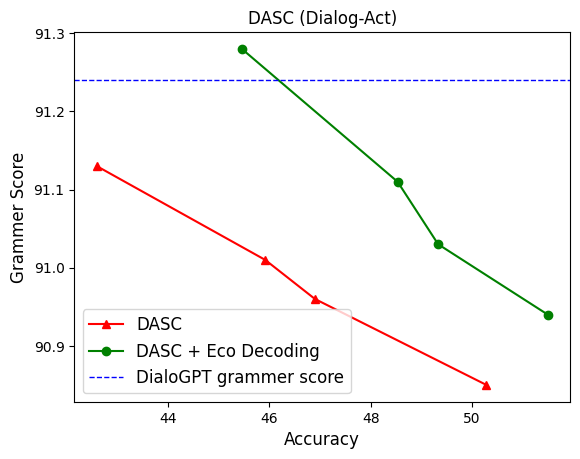}  
    \caption{The single attribute control performance of the existing weighted decoding method (red) and ECO decoding (green) with respect to changes in the control strength $\lambda$. The y-axis represents grammar, and the x-axis represents accuracy. The blue dot line represents uncontrolled DialoGPT's grammar score.}
    \label{fig:single}
\end{figure}

\begin{figure}[t!]
\vspace{-1mm}
    \centering
    \includegraphics[width=0.235\textwidth]{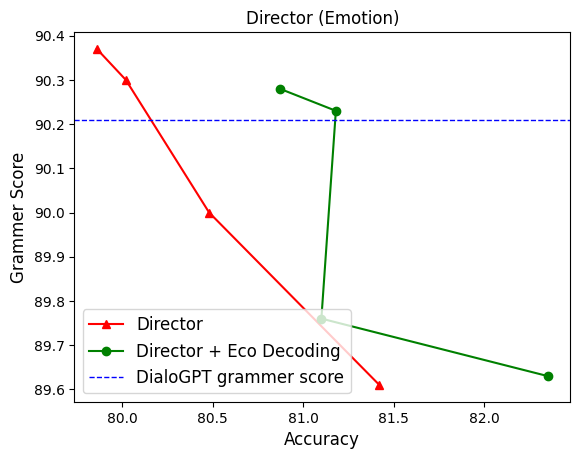}
    \includegraphics[width=.235\textwidth]{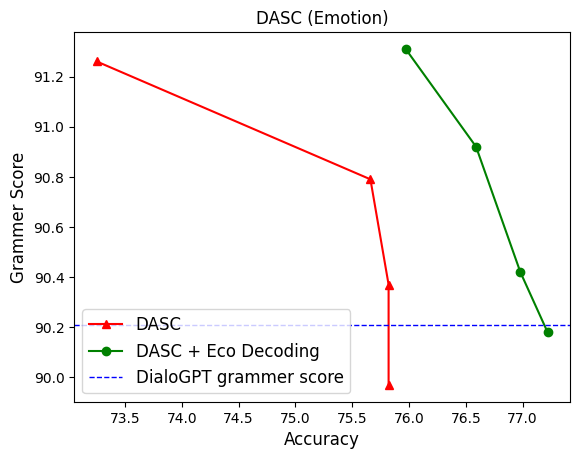}

    \textbf{\includegraphics[width=.235\textwidth]{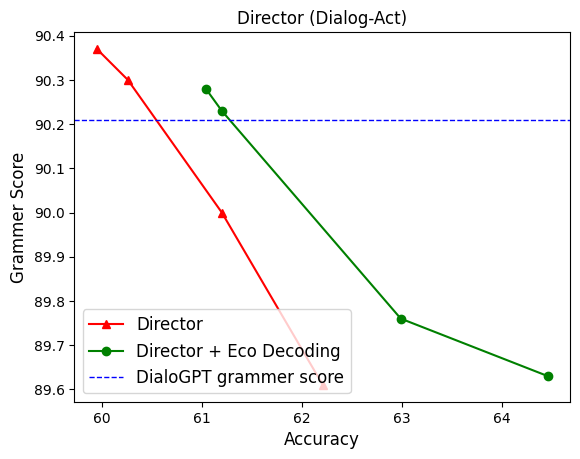}}
    \includegraphics[width=.235\textwidth]{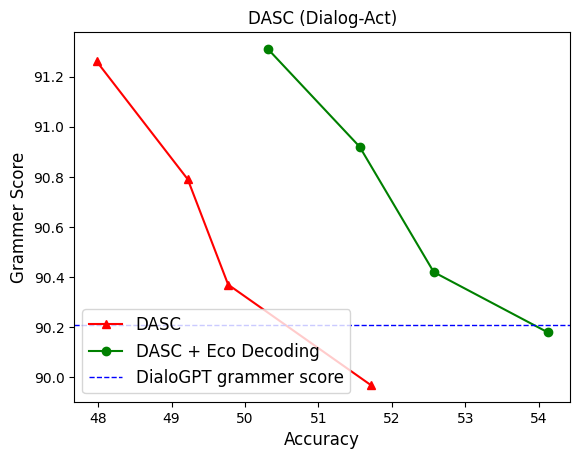}  
    \caption{The multi attribute control performance of the existing weighted decoding method (red) and ECO decoding (green) with respect to changes in the control strength $\lambda$. The y-axis represents grammar, and the x-axis represents accuracy. The blue dot line represents uncontrolled DialoGPT's grammar score.}
    \label{fig:multi}
\end{figure}

\paragraph{Robustness Test}
The control strength coefficient $\lambda$ determines the proportion of weights in the probability distribution of the attribute classifier. Therefore, a larger lambda will tend to generate tokens that are more attribute-specific, resulting in a trade-off between increased attribute accuracy and decreased grammar score.

Figure \ref{fig:single} shows the results of applying traditional weighted decoding methodology and ECO-decoding for varying the control strength coefficient $\lambda$ in a single attribute control setting. The experiments were conducted based on the Director and DASC method for two attributes, emotion and dialog-act, and the red line is for the Director and DASC. The green line is for the application of ECO decoding with each method and the blue bashed line is the grammar score of vanila DialoGPT without attribute control according to each dataset.
In all experiments, we observed a tradeoff between grammar and accuracy, and showed that for the same grammar score, ECO decoding achieves higher attribute accuracy by dynamically applying weights. In other words, for the same control degree, ECO decoding produces higher quality responses.
This demonstrates that our approach has a strong capability in controllable generation to maintain fluency while enhancing controllability, regardless of the $\lambda$ values.

Figure \ref{fig:multi} shows the performance by lambda in the multi attribute control setting. The results are the same for multi attribute as for single, with ECO decoding for each methodology resulting in higher controllability and grammar scores. The interesting thing is that the Director model is not a structured model for multi attribute control, and because of this, there is some performance variation by lambda.

In almost all cases using the DialoGPT-small model, we observed a trend of grammar score increase and then decrease as the lambda value increases. Due to the increase in grammaticality at low $\lambda$ ranges, some experimental results showed higher grammar score with attribute control than DialoGPT without attribute control. This was not observed when using larger models such as Llama2, which could be thought of as a slight inconsistency in the performance of small language models. For fairness, all experimental results measured grammar score and attribute accuracy after the lambda value where the trade-off occurs.

\paragraph{Stability of Entropy Smoothing}
To examine the impact of entropy smoothing, we varied the temperature parameter $\tau$ in the softmax normalization process and measured grammar quality and control accuracy for the Emotion attribute on the DailyDialog dataset using the DialoGPT (176M) model with the Director method, with $\tau$ values ranging from 0.1 to 10.

\begin{table}[t]
\centering
\resizebox{0.8\linewidth}{!}{
\begin{tabular}{ccc}
\toprule
\textbf{Temperature ($\tau$)} & \textbf{Grammar} & \textbf{Accuracy} \\
\midrule
0.1  & 90.30 & 82.74 \\
0.5  & 90.30 & 82.82 \\
1.0  & 90.30 & 82.82 \\
5.0  & 90.28 & 83.28 \\
10.0 & 89.57 & 85.69 \\
\bottomrule
\end{tabular}
}
\caption{Effect of varying the entropy smoothing temperature $\tau$ on grammar and control accuracy.}
\label{tab:entropy_smoothing}
\end{table}

As shown in Table~\ref{tab:entropy_smoothing}, performance remained stable when $\tau$ was varied between 0.1 and 5.0, with only minor fluctuations in grammar and accuracy. At an excessively large value ($\tau = 10$), some deviations were observed, suggesting that heavy smoothing may reduce the ability of entropy to capture model uncertainty and affect the behavior of dynamic control. Overall, the results indicate that ECO decoding is not overly sensitive to reasonable choices of the smoothing parameter.

\section{Conclusions}
We propose ECO decoding as an entropy-based approach for dynamically adjusting control strength at each step in weighted decoding. Unlike prior methods relying on static coefficients, ECO decoding utilizes the uncertainty of each attribute classifier during inference to improve controllability without sacrificing fluency.

ECO decoding is easily applicable to existing controllable generation methods, as it requires no additional training. It effectively generalizes across single- and multi-attribute settings, and properly addresses interpolation issues in multi-attribute control by leveraging attribute-specific entropy.

Both automatic and human evaluations demonstrate that ECO decoding consistently outperforms static baselines across a range of models and control strengths, including strong performance on large-scale models like Llama2-7B. These findings highlight that ECO decoding is an effective solution for more precise and robust controllable dialogue generation.

\section*{Limitations}

Our method improves control performance while maintaining sentence fluency by leveraging entropy-based control strength, and it enables fluent sentence generation in both single- and multi-attribute settings. However, this approach has several limitations. First, the dataset used in this study is limited in both quantity and diversity, and validation was performed only on dialogues with two attributes. Therefore, construction of new corpus with additional attributes is necessary to evaluate the generalizability of the proposed method. Second, in multi-attribute settings, a refined normalization process must be developed to account for the number of classes for each attribute as the number of attribute classifiers increases. To address these limitations, future work should construct more diverse datasets and explore effective probability control strategies under multi-attribute conditions.

\section*{Ethics Statement}
The proposed method aims to enhance the interest and accuracy of responses generated by chatbots to improve user experience. However, this method could be potentially used for malicious purposes. In our experiments, we focus on attributes like emotion and dialog-act, but if malicious desired attributes such as bias are used, the model could be induced to generate inappropriate responses. Therefore, generating controlled responses using malicious attributes should be restricted.

\section*{Acknowledgments}
This work was partly supported by the National Research Foundation of Korea (NRF) grant funded by the Korea government (MSIT) (No. RS-2024-00350379, 30),  Institute for Information \& communications Technology Promotion(IITP) grant funded by the Korea government(MSIT)(RS-2024-00343989, Enhancing the Ethics of Data Characteristics and Generation AI Models for Social and Ethical Learning, 30), Institute of Information \& Communications Technology Planning \& Evaluation(IITP)-ITRC(Information Technology Research Center) grant funded by the Korea government(MSIT) (IITP-2025-RS-2024-00437633, 20), and Institute of Information \& communications Technology Planning \& Evaluation(IITP) grant funded by the Korea government(MSIT) (RS-2019-II190421, Artificial Intelligence Graduate School Program(Sungkyunkwan University), 20)

\bibliography{custom}

\appendix

\section{Analysis}
\subsection{POS-Based Entropy Analysis}
To investigate the theoretical foundation of entropy-aware control strength, we conducted a token-level entropy analysis based on part-of-speech (POS) tags. Entropy scores are measured using the token probability distributions from the DialoGPT models.

In Table \ref{tab:Anaylsis_pos_entropy}, we can observe some patterns in the relationship between a part-of-speech tag and its entropy. Functional words such as ADP (adpositions), DET (determiners), and PUNCT (punctuation) tend to have relatively low entropy because they are typically governed by strict grammatical constraints that limit lexical variability during generation. In contrast, content words including NOUN (nouns), VERB (verbs), ADJ (adjectives), and ADV (adverbs) exhibit higher entropy because these words often occur in more flexible contexts and are responsible for conveying core semantic information, which results in a broader distribution over candidate tokens.

These findings provide a theoretical support for our entropy-based control strategy. Specifically, positions with high entropy exhibit greater lexical variability, which allows for more flexible attribute manipulation without compromising the fluency of the generated text. In contrast, low-entropy tokens are typically constrained by grammatical structure and should therefore be subjected to weaker control in order to maintain grammatical correctness. This POS-tag-based analysis confirms that entropy can serve as an effective indicator for determining the appropriate timing and location of control application during the generation process.
\begin{table}[]
\centering
\resizebox{0.9\columnwidth}{!}{%

\begin{tabular}{lcc}
\hline
\textbf{POS Tag} & \textbf{DialoGPT-S} & \textbf{DialoGPT-L} \\
\hline\hline
ADP   & 3.04 & 3.19 \\ \hline
CCONJ & 3.68 & 4.04 \\\hline
PUNCT & 3.69 & 3.75 \\\hline
PART  & 3.78 & 3.51 \\\hline
SCONJ & 4.20 & 4.37 \\\hline
DET   & 4.27 & 4.26 \\\hline
PRON  & 4.35 & 4.40 \\\hline
ADV   & 4.41 & 4.55 \\\hline
NUM   & 4.51 & 4.71 \\\hline
X     & 4.55 & 4.94 \\\hline
VERB  & 4.58 & 4.74 \\\hline
NOUN  & 4.81 & 5.39 \\\hline
SYM   & 4.85 & 5.31 \\\hline
PROPN & 4.85 & 5.29 \\\hline
ADJ   & 5.00 & 5.35 \\\hline
AUX   & 5.10 & 4.93 \\\hline
INTJ  & 5.47 & 5.71 \\\hline
\end{tabular}
}
\caption{Entropy measurement score in each part-of-speech (POS) tag. Results are reported for the DialoGPT models: S (small) and L (large). Items in Table 7 are sorted in ascending order based on the entropy scores of DialoGPT-S.}
\label{tab:Anaylsis_pos_entropy}
\end{table}

\begin{table*}[t!]
\centering
{
\resizebox{1\textwidth}{!}{ 
\begin{tabular}{lccccccc}
\hline
Model & Accuracy(emo) & Accuracy(act) & Rouge-1 & Rouge-L & Dist-1 & Dist-2 & Grammar \\
\hline \hline
DialoGPT & \multicolumn{1}{c}{-} & \multicolumn{1}{c}{-} & 9.00 & 8.53 & 0.58 & 0.76 & 90.21 \\
Director & 80.48 & 60.65 & 9.41 & 8.99 & 0.58 & 0.73 & 90.22 \\
\texttt{- lm\_entropy} & 79.24 & 58.63 & 9.49 & 9.04 & 0.58 & 0.73 & \textbf{90.31} \\
\texttt{- exponential} & 80.48 & 61.12 & 9.42 & 9.01 & 0.58 & 0.73 & 90.21 \\
\texttt{- negative} & 80.17 & 60.96 & 9.41 & 9.00 & 0.58 & 0.73 & 90.21 \\
\texttt{- reciprocal} & \textbf{81.18} & \textbf{61.20} & 9.49 & 8.97 & 0.58 & 0.73 & 90.23 \\
\hline
\end{tabular}
}
}
\caption{Experiments on control strength application methods. We compare the performance of the lm\_entropy method, which utilizes the probability distribution of the language model, with three alternative methods that use the probability distribution of the attribute classifier.}
\label{tab:analysis_control_strength}
\end{table*}

\begin{table*}[t!]
\centering
\resizebox{1\textwidth}{!}{ 
\begin{tabular}{lccccccc}
\hline
Attribute & Accuracy(emo) & Accuracy(act) & Rouge-1 & Rouge-L & Dist-1 & Dist-2 & Grammar \\  
\hline \hline
Emotion     & 76.28 & \multicolumn{1}{c}{-}    & 15.55 & 13.34 & 0.34 & 0.80 & 90.10 \\  
Dialog-act  & \multicolumn{1}{c}{-}    & 40.12 & 15.19 & 12.87 & 0.33 & 0.79 & 89.59 \\  
Multi       & 58.71 & 29.62 &  6.24 &  5.52 & 0.26 & 0.55 & 85.22 \\  
\hline
\multicolumn{8}{c}{\textbf{Prompt}} \\ \hline
\multicolumn{8}{c}{
\begin{minipage}{0.9\textwidth}\centering
You are a dialogue system that engages in everyday conversation with a user.\par
Below is a conversation history consisting of multiple turns, with each turn labeled as "User:" or "System:" to indicate the user's utterance and the system's response, respectively.\par

You must generate a response that satisfies the condition: \{label\}: \{attribute value\}.\par

Based on the provided conversation history, generate the next system response that continues the conversation naturally.
\end{minipage}
} \\  
\hline
\end{tabular}
}
\caption{Zero-shot prompting results with Llama2 on the DailyDialog test set.}
\label{tab:zeroshot}
\end{table*}

\subsection{Control Strength Analysis}
In this study, to mathematically model control strength as inversely proportional to the entropy of the token distribution, we define control strength as a function of entropy and employ three representative decreasing functions. These functions are designed to dynamically adjust control strength depending on the level of entropy, thereby enabling more assertive control in confident contexts (i.e., low entropy) and weaker control when the model is uncertain (i.e., high entropy).

The three control weighting functions are as follows:
\vspace{-2mm}
\begin{equation}
\label{eq:exponential}
exponential = 1 + exp(-attr\_entropy)
\end{equation}

\vspace{-2mm}
\begin{equation}
\label{eq:negative}
negative = 1 + (log(V) - attr\_entropy)
\end{equation}

\vspace{-2mm}
\begin{equation}
\label{eq:inverse}
reciprocal = 1 + (\frac{1}{attr\_entropy})
\end{equation}

\begin{table}[t]
\centering
\resizebox{0.9\linewidth}{!}{
\begin{tabular}{lcccc}
\toprule
\textbf{Method} & \textbf{Static Control} & \textbf{ECO decoding} & \textbf{Overhead} & \textbf{Increase} \\
\midrule
FUDGE    & 3.087 ms & 3.114 ms & $\times$1.009 & +0.9\% \\
Director & 1.143 ms & 1.160 ms & $\times$1.015 & +1.5\% \\
DASC     & 1.116 ms & 1.123 ms & $\times$1.006 & +0.6\% \\
\bottomrule
\end{tabular}
}
\caption{Average per-token decoding latency on the DailyDialog dataset with emotion control.}
\label{tab:latency}
\end{table}

When comparing the three functions, the linear function maintains a direct linear relationship between entropy and weight. In contrast, both the exponential function  and the inverse function  yield similar weights in high-entropy regions, but diverge significantly in low-entropy regions with assigning notably larger weights.

Table \ref{tab:analysis_control_strength} shows that the linear function tends to degrade performance in multi-attribute control tasks. This is due to the fact that its weights decrease linearly without convergence as entropy increases, resulting in unstable control. On the other hand, both the exponential and reciprocal functions exhibit convergence as entropy increases, and outperform the baseline model. These two functions produce larger differences in weights in low-entropy regions, and our experiments confirm that the reciprocal function particularly achieves strong performance. This is more likely because, in cases where the probability distribution is highly concentrated (i.e., one token is assigned a very high probability), the reciprocal function is better suited to apply strong control.

Furthermore, the reciprocal function provides more advantages than the exponential function by offering a broader range of control strength in low-entropy settings. Consequently, we conclude that the reciprocal weighting function is well-suited for assigning control strength appropriately under both low-entropy and high-entropy conditions.

\section{Zero-shot Prompting with Llama2}
We additionally examined zero-shot prompting as an alternative approach to attribute control. 
All experiments were conducted on the DailyDialog test set under single-attribute (emotion, dialog-act) and multi-attribute control settings. 
For zero-shot prompting, we used Llama2 with task-specific instructions, without any fine-tuning or decoding-based control.

As shown in Table~\ref{tab:zeroshot}, zero-shot prompting can control certain attributes to some extent, but its overall controllability remains limited. Compared to DialoGPT (176M), Llama2 with zero-shot prompting, despite being a larger model, showed lower performance than Director and FUDGE in emotion control. When compared with DASC, it achieved higher controllability but at the cost of reduced grammaticality. For dialog-act and multi-attribute control, its performance was substantially lower than decoding-based approaches in both controllability and grammaticality. These results suggest that while prompt-based approaches offer some degree of controllability, decoding-based methods remain more useful for achieving a balance between controllability and fluency.

\paragraph{Impact of Entropy Calculation on Inference Speed}
To assess the inference-time overhead of ECO decoding, we measure the average per-token generation time on the DailyDialog dataset with emotion control using the DialoGPT (176M) model. We compared ECO decoding against static control across three weighted decoding baselines (FUDGE, Director and DASC). Despite requiring entropy computation for both the language model and the attribute classifier at every decoding step, ECO decoding introduced only minimal additional latency.

As shown in Table~\ref{tab:latency}, the decoding overhead of ECO decoding was less than 1.5\% across all baselines. Because our method only requires entropy computation on top of existing weighted decoding operations, the additional cost is negligible. These results indicate that ECO decoding is well suited for real-time inference scenarios, achieving controllability improvements while maintaining efficient generation speed.

\section{Cases of ECO decoding}
\begin{figure}[h]
\vspace{-1mm}
    \centering
    \includegraphics[width=.5\textwidth]{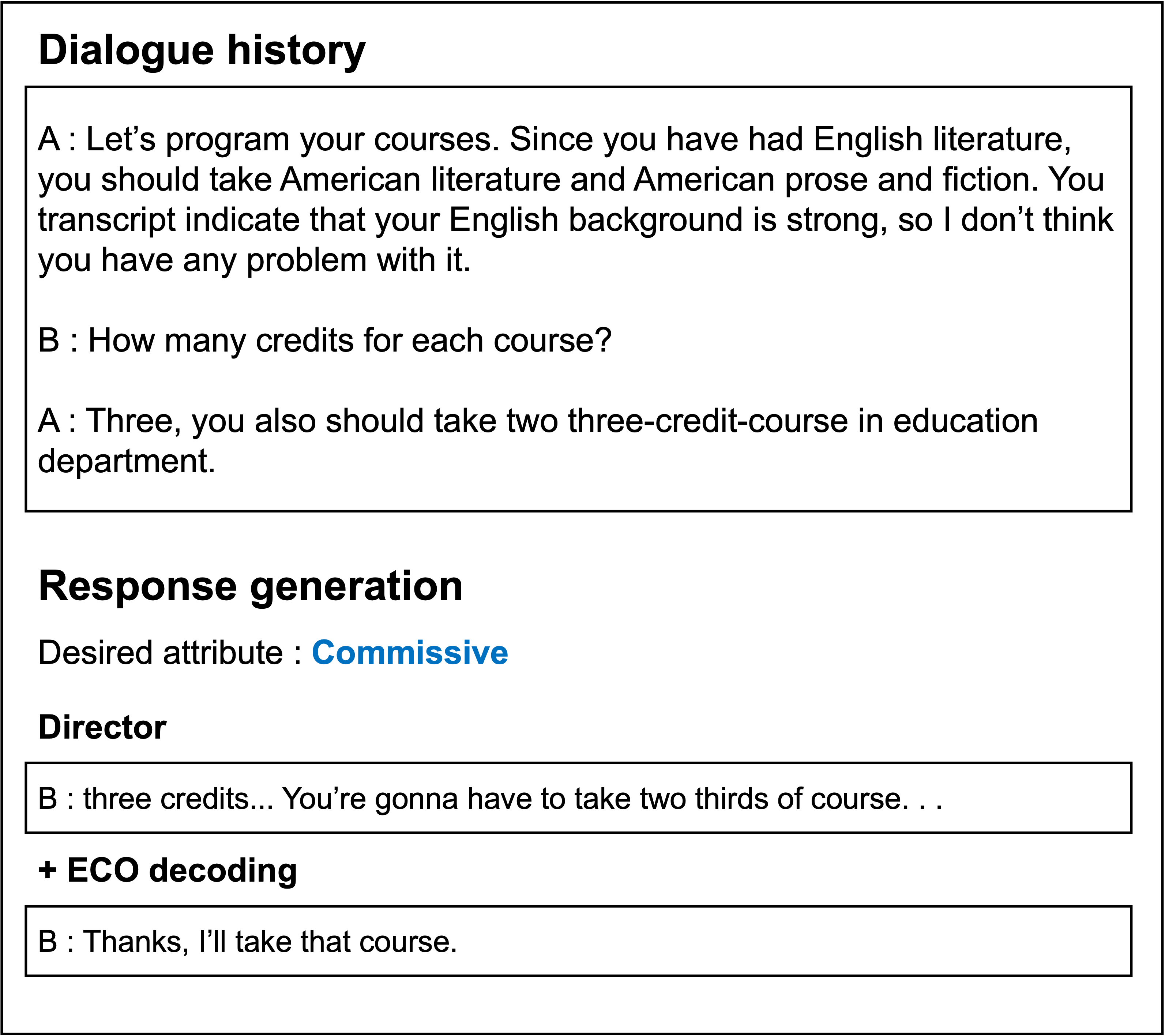} 
    \caption{In case where the response fails to satisfy the desired attribute with the existing method but satisfies the desired attribute using ECO decoding.}
    \label{fig:appendix1}
\end{figure}
\begin{figure}[h]
\vspace{-1mm}
    \centering
    \includegraphics[width=.5\textwidth]{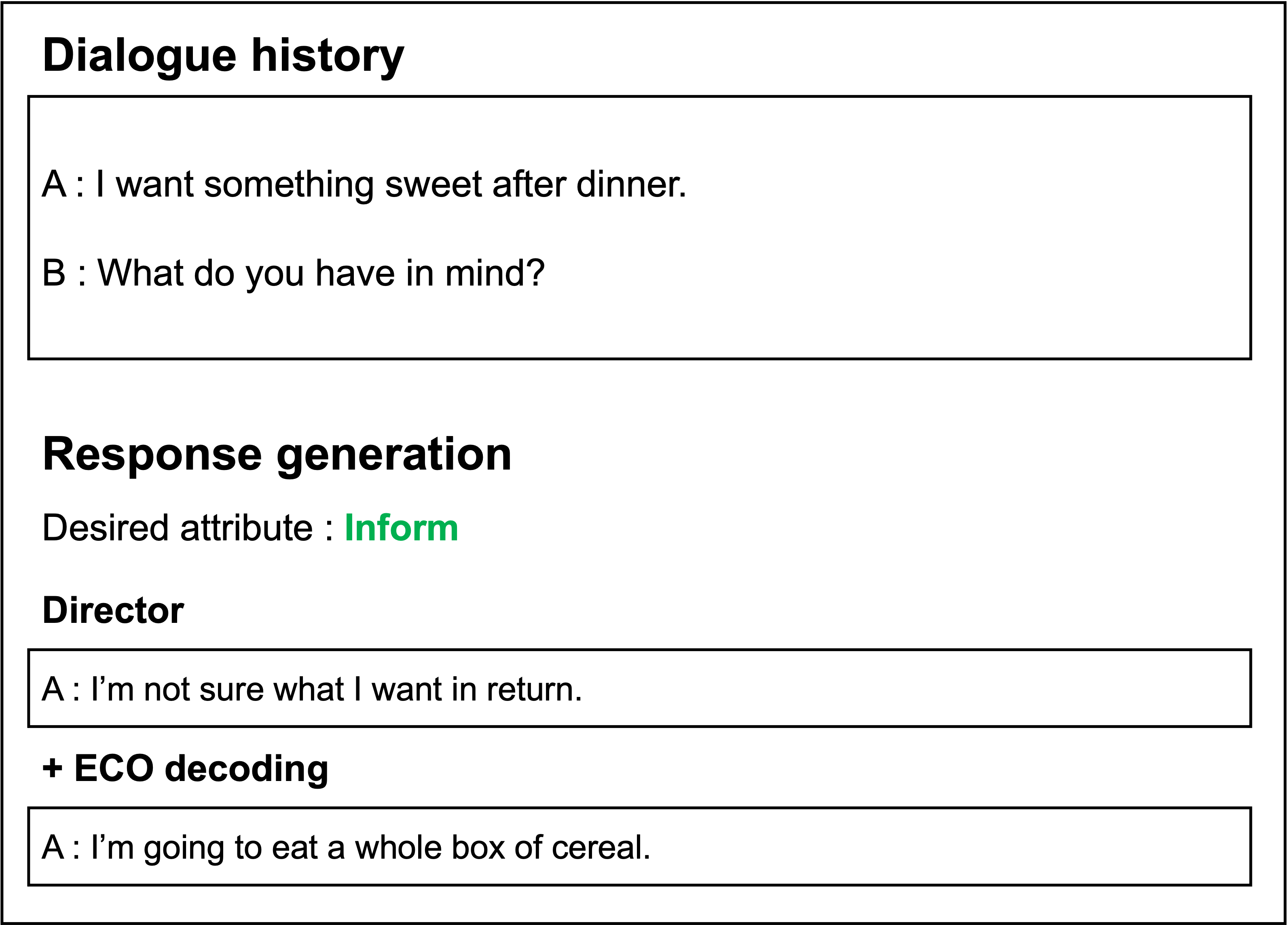} 
    \caption{
    In case where it fails to generate a context-consistent response with the existing method but generates a context-consistent response using ECO decoding.}
    \label{fig:appendix2}
\end{figure}

\section{Licenses}
The DailyDialog dataset is licensed under CC BY-NC-SA 4.0 License.
The DialoGPT model is licensed under Contributor License Agreement (CLA) and Llama2 model is licensed under Meta Llama 2 Community License Agreement.
The RoBERTa-based CoLA grammaticality model is licenced under MIT License.

\end{document}